\pdfoutput=1

\documentclass[11pt]{article}

\usepackage[]{acl}

\usepackage{times}
\usepackage{latexsym}
\usepackage{pifont}
\usepackage{booktabs}
\usepackage{multirow}
\usepackage{graphicx}

\newcommand{\cmark}{\ding{51}}%
\newcommand{\xmark}{\ding{55}}%

\usepackage[T1]{fontenc}

\usepackage[utf8]{inputenc}

\usepackage{microtype}

\title{SemEval-2022 Task 7: Identifying Plausible Clarifications of\\ 
Implicit and Underspecified Phrases in Instructional Texts}

\author{Michael Roth \quad\quad Talita Anthonio
\quad\quad Anna Sauer 
\\
University of Stuttgart \\
Institute for Natural Language Processing \\
\small{\texttt{\{michael.roth,talita.anthonio,anna.sauer\}@ims.uni-stuttgart.de}}}

\begin{document}
\maketitle

\newcommand\ecfootnote[1]{%
\begingroup
\renewcommand\thefootnote{}\footnotetext{#1}%
\addtocounter{footnote}{-1}%
\endgroup
}

\begin{abstract}
We describe SemEval-2022 Task 7, a shared task on rating the plausibility of clarifications in English-language instructional texts. 
The dataset for this task consists of manually clarified how-to guides for which we generated alternative clarifications and collected human plausibility judgements.\footnote{The task data is available at \url{https://github.com/acidAnn/claire}.} The task of participating systems was to automatically determine the plausibility of a clarification in the respective context. In total, 21 participants took part in this task, with the best system achieving an accuracy of 68.9\%. This report summarizes the results and findings from 8 teams and their system descriptions. Finally, we show in an additional evaluation that predictions by the top participating team make it possible to identify contexts with multiple plausible clarifications with an accuracy of 75.2\%. 
\end{abstract}

\section{Introduction}
\label{sec:intro}

Understanding texts in natural language requires that both explicit text components as well as implicit references and relationships are interpreted correctly. This applies in particular to instructional texts, which demand a clear understanding of individual instruction steps in order to reach the desired goal. Possible uncertainties should therefore already be clarified in the text. In principle, such clarifications can also be generated automatically. In that case, however, it will be necessary to investigate the circumstances under which a clarification is plausible and unambiguous.

As a first step towards such an investigation, this shared task evaluates the ability of NLP systems to distinguish between plausible and implausible clarifications of an instruction. 
Inspired by the success of previous cloze-based evaluations (see Section~\ref{sec:relwork}), we set up our task as a cloze task, in which clarifications are presented as fillers and systems have to identify which fillers plausibly fit in a given context (see Table~\ref{tbl:example}). Our focus in this task is on different types of referring expressions that are either underspecified or not realized explicitly at all, and we consider possible clarifications in the form of additional specification or explicitation.  

Research in linguistics and psychology has shown that individuals use language differently \cite{Pennebaker:1999:styles,Heylighen:2002:variation}. In particular when it comes to implicit and underspecified language, individual differences can also lead to different interpretations \cite{scholman-demberg-2017-crowdsourcing,poesio-etal-2019-crowdsourced}. 
As a result, worst case scenarios include medical instructions being followed incorrectly or news being passed on inaccurately.
In view of the fact that language is inherently ambiguous, however, it is neither sensible nor expedient to produce clarifications for all occurrences of underspecification. Avoiding worst-case scenarios therefore goes beyond ranking individual clarifications by plausibility and must take into account whether multiple (incompatible) clarifications are perceived as plausible, thus reflecting possible misunderstandings. 

\begin{table}[t!]
\centering
\begin{tabular}{cl@{\quad\quad}cl}
\toprule
\multicolumn{4}{c}{\textbf{Choose a Hair Salon}} \\
\midrule
(1) & \multicolumn{3}{l}{Check ratings of different salons.}\\
(2) & \multicolumn{3}{l}{Visit the salon's website.}\\
(3) & \multicolumn{3}{l}{Call \underline{~~~~$\emptyset$~~~~} and ask questions.}\\
\midrule
\cmark & the salon  & \xmark & a friend\\
\cmark & the number & \xmark & your stylist\\
\cmark & \multicolumn{3}{l}{the owner}\\
\bottomrule
\end{tabular}
\caption{Simplified example from a pilot study: the top shows a sentence (3) and shortened version of its discourse context (1--2). In the clarified version of this sentence, the phrase \textit{the salon} was inserted. Other phrases shown in the bottom part are automatically generated fillers, annotated as plausible (\cmark) or implausible (\xmark).}
\label{tbl:example}
\end{table}

We discuss our task and data in more detail in Sections~\ref{sec:task} and~\ref{sec:data}, respectively. The participating teams are summarized in Section~\ref{sec:participants} and their results on the task and additional evaluations in Section~\ref{sec:results}.

\section{Related Work}
\label{sec:relwork}

Cloze tasks have become a standard framework for evaluating various discourse-level phenomena in NLP. Some prominent examples include the narrative cloze test \cite{chambers-jurafsky-2008-unsupervised}, the story cloze test \cite{mostafazadeh-etal-2016-corpus}, and the LAMBADA word prediction task \cite{paperno-etal-2016-lambada}. In these tasks, NLP systems are required to make a prediction about the filler of a cloze that is most likely to continue the discourse. 
However, it is not always clear whether exactly one likely filler exists. Evaluations typically circumvent this issue by requiring systems only to distinguish between a correct and an incorrect filler, or by evaluating predictions only with a relative measure. Both of these options ignore the more general challenge that multiple fillers can be plausible. 
Our shared task addresses this challenge explicitly, by requiring systems to classify different clarifying fillers as either plausible or implausible. This is a natural extension of previous cloze tasks to discourse contexts in which multiple interpretations are plausible. This extension makes it possible to evaluate in how far NLP systems can reflect cases of underspecification and uncertainty as well as possible sources of misunderstanding.

Our task is based on manual text revisions that can be traced through revision histories and by which possible needs for clarification can be identified. Thus, the task follows a number of existing research contributions that deal with text revisions. A number of previous works examine reasons for and types of revisions in, for example, Wikipedia \cite{bronner-monz-2012-user,daxenberger-gurevych-2012-corpus,yang-etal-2017-identifying-semantic,faruqui-etal-2018-wikiatomicedits} and the essay-based corpus ArgRewrite \cite{zhang-litman-2015-annotation,zhang-litman-2016-using,afrin:2019:identifying,kashefi:2022:argrewriteV2}. In this work, we use revision histories of instructional texts because clarifications seem particularly relevant in this domain.

The starting point of our task is the data set wikiHowToImprove \cite{anthonio-etal-2020-wikihowtoimprove}, which comprises revision histories of more than 250,000 how-to guides from the online platform wikiHow.\footnote{\texttt{www.wikihow.org}} In our own previous work, we  investigated the extent to which these histories are useful for examining textual improvements \cite{anthonio-roth-2020-learn}, predicting revision requirements \cite{bhat-etal-2020-towards}, modeling cases of lexical vagueness \cite{debnath-roth-2021-computational}, and resolving implicit references \cite{anthonio-roth-2021-resolving}. The last two studies in particular have shown that wikiHowToImprove is well suited as a resource for studying clarifications of semantic phenomena. We describe one of these studies and how we build upon it in more detail in Section~\ref{sec:task}.

\section{Task and Background}
\label{sec:task}

The general idea of the present shared task is to use revisions of English-language instructional texts as a basis to identify potential clarifications and to rate them regarding their plausibility. We assume that at least certain cases of clarifying revisions follow patterns that can be recognized automatically, by comparing the text before/after revision. An example of such a pattern is the insertion of a nominal phrase mentioned in context that makes an implicit reference explicit (see Table~\ref{tbl:example}).
We consider additional patterns as part of this shared task (see Section~\ref{sec:data}), but only consider cases of insertion for simplicity. 

The focus on insertions allows us to consider clarifications as solutions to a cloze test, since the revision always fills in a text segment that previously was not present. Compared to previous cloze tasks, we do not assume that the revision observed is always unique and plausible. Instead, we also consider alternative clarification options and obtain plausibility judgments for all options.

As background, we first summarize findings from a pilot study that we conducted before setting up the present task (\S\ref{subsec:pilot}). Based on this, we then describe the settings of the shared task (\S\ref{sec:eval}).

\subsection{Pilot Study}
\label{subsec:pilot}

In our pilot study \cite{anthonio-roth-2021-resolving}, we constructed a dataset of implicit references and potential clarifications in three steps: (1) heuristically identifying insertions of nominal phrases mentioned in the previous context, (2) automatically generating alternative clarifications using generative language modeling \cite[GPT;][]{Radford:2018:gpt}, and (3) collecting human plausibility judgements for each clarification option. 

The first step of our pilot showed that it is possible to extract about 6,000 relevant clarifications from the revisions in wikiHowToImprove. We further found that most noisy instances can be filtered by the application of linguistic constraints and that remaining cases can be identified during the manual verification in the final step. In the second step, we found GPT to produce completions for many sentences that seem sensible on the surface level. Using our best strategy, namely re-ranking based on paragraph-level perplexity, the best sequence generated by the model was identical to the human-inserted clarification in over 56\% of cases, and the clarification appeared in the top-10 generated sequences in 78\% of cases.

A crucial finding in the third step was that the annotator indicated a preference for the human-inserted clarification in most cases (68\%), but different, model-generated clarifications were judged as equally good in many cases (24\%). In some cases, the annotator actually preferred a generated clarification over the human-edited insertion (8\%).

The framework for the shared task is strongly motivated by the finding that alternative clarifications, generated by a computational model, can be as good or even better than human-produced clarifications. In some cases, we simply found different verbalizations of the same proposition. In other cases, like examples (a) and (b) below, we found plausible alternatives that are not fully compatible semantically.

\begin{itemize}
\item[(a)] Call \underline{the salon} and ask questions.
\item[(b)] Call \underline{each salon} and ask questions.
\end{itemize}

When multiple incompatible readings exist, there is a risk that instructions will be misunderstood and not lead to the desired goal. To identify potential occurrences of such cases, we consider different fillers in the shared task and rate the plausibility of each filler independently.

\begin{table*}[!ht]
    \centering
    \begin{tabular}{@{~}llll@{~}}
    \toprule
    Phenomenon & Clarification pattern & Example & Potential filler\\
    \midrule
    \multirow{5}{2cm}{Implicit Reference}  & \multirow{4.5}{5cm}{$\emptyset$ $\rightarrow$ [DET] NOUN} &
    \multirow{5}{5cm}{Lift your toes up while keeping your leg straight. Hold \underline{~~~~$\emptyset$~~~~} for a few seconds, then release. Incorporate calf stretches into your yoga routine.} 
    & \cmark ~your pose \\
    & & & \cmark ~the stretch\\
    & & & $\textbf{?}$ a leg\\
    & & & $\textbf{?}$ the chair\\
    & & & \xmark ~your head\\
    \midrule
    \multirow{5}{2cm}{Fused head}  & \multirow{5}{4.5cm}{DET/JJ $\emptyset$ $\rightarrow$ DET/JJ NOUN} &
    \multirow{5}{5cm}{Traditionally, the groom waits for the bride at the altar, the bride tosses the bouquet and (\ldots). Since this is your wedding, feel free to change these \underline{~~~~$\emptyset$~~~~}.} 
    & \cmark ~ideas \\
    & & & \cmark ~plans\\
    & & & $\textbf{?}$ symbols\\
    & & & $\textbf{?}$ characters\\
    & & & \xmark ~changes\\
    \midrule
    \multirow{5}{2cm}{Noun compound}  & \multirow{5}{4.5cm}{$\emptyset$ NOUN $\rightarrow$ NOUN NOUN} &
    \multirow{5}{5cm}{Heating for cold water tanks isn't quite of an issue as for tropicals. In fact, you can keep a \underline{~~~~$\emptyset$~~~~} tank without a heater.} 
    & \cmark ~goldfish  \\
    & & & \cmark ~freshwater \\
    & & & \cmark ~water\\
    & & & $\textbf{?}$ fishing \\
    & & & \xmark ~soup\\
    \midrule
    \multirow{5}{2cm}{Metonymy}  & \multirow{5}{4.5cm}{NP $\emptyset$ $\rightarrow$ NP's NP\\ $\emptyset$ NP $\rightarrow$ NP of NP} &
    \multirow{5}{5cm}{Look at the \underline{~~~~$\emptyset$~~~~} of the teeth. If you're unsure of your dog's age, or want to determine if they are already entering into the senior territory, try the teeth.} 
    & \cmark ~condition \\
    & & & \cmark ~color \\
    & & & \cmark ~thickness \\
    & & & $\textbf{?}$ layout\\
    & & & $\textbf{?}$ points\\
    \bottomrule
    \end{tabular}
    \caption{Phenomena, 
    extraction patterns 
    and example 
    clarifications (\cmark ~plausible, $\textbf{?}$ neutral, \xmark ~implausible).
    }
    \label{tbl:aspects}
\end{table*}

\subsection{Shared Task Settings}
\label{sec:eval}

The SemEval shared task is set up as follows: Systems are provided with a cloze sentence, surrounding sentences and a potential clarifying filler as input, and are required to make a prediction regarding the plausibility of the filler in the given context. For evaluation, predicted labels are compared against the manually collected plausibility judgements described in Section~\ref{sec:data}. We define two subtasks with different labels and evaluation measures.

\paragraph{Task~1: Classification.} In the classification task, systems need to distinguish between three labels (\textsc{implausible}, \textsc{neutral} and \textsc{plausible}). We use \textit{accuracy} as the main evaluation measure, calculated as the proportion of correct predictions among all predictions of a system.

\paragraph{Task~2: Ranking.} In the ranking task, systems need to predict a continuous plausibility score. We evaluate the predictions based on their \textit{correlation} with human judgements, calculated as Spearman's rank correlation coefficient between all predictions and all judgements.

We describe the selection of data and collection of human judgements in the next section. In Section~\ref{sec:results}, we discuss additional evaluations performed to assess system performance with regard to the presence of multiple plausible clarifications.

\section{Data}
\label{sec:data}

We closely follow the three steps of our pilot study, described in Section~\ref{subsec:pilot}, to construct the data for this shared task. Our starting point is the dataset wikiHowToImprove \cite{anthonio-etal-2020-wikihowtoimprove}, a resource of sentence-level revisions and their contexts based on wikiHow. 
In the first two steps, we create relevant data from this resource automatically; in the final step, we collect manual annotations to form a gold standard. In step 1, we apply a pattern-based approach to identify revisions that involve insertions that serve specific clarifying functions (\S\ref{subsec:step1}). In step 2, we use transformer-based language models to produce sets of alternate clarifications that may or may not be compatible with an observed insertion (\S\ref{subsec:step2}). In step 3, we collect human plausibility judgements on each clarification independently (\S\ref{subsec:step3}).

\subsection{Data Extraction}
\label{subsec:step1}

We collect relevant revisions by identifying cases in which a single \textit{contiguous} insertion and no other change was made within a sentence. We compute differences and extract cases automatically based on the Python library \texttt{difflib}\footnote{\url{https://docs.python.org/3/library/difflib.html}}
and the following preprocessing tools:
\texttt{spaCy}\footnote{\url{https://github.com/explosion/spaCy}} for sentence splitting and tokenization, the Berkeley Neural Parser \cite{kitaev-klein-2018-constituency}
for constituency parsing and \texttt{Stanza} \cite{qi-etal-2020-stanza} for POS tagging, dependency parsing and co-reference resolution.
For the shared task, we focus on four types of phenomena, which are summarized in Table~\ref{tbl:aspects}.

\paragraph{Implicit references.} Instances with a non-verbalized reference in the original sentence which was clarified in the revised sentence through insertion. We select the cases from \newcite{anthonio-roth-2021-resolving} with insertions containing a single noun or a determiner followed by a noun. 

\paragraph{Fused heads.} Instances of noun phrases for which the head noun was implicit in the original sentence and clarified in the revised sentence through insertion. We search for noun phrases with a determiner or adjective head in the original sentence and select those instances where a single noun was inserted in the revision.
\paragraph{Noun compounds.} Instances of underspecified noun phrases, which were clarified in the revised sentence through the insertion of a dependent noun to form a more specific compound. We select instances of single noun insertions in which the inserted noun is a \texttt{compound} dependent of another noun that has already been present in the original sentence.

\paragraph{Metonymy.} Instances in which a revision adds a noun $y$ to a noun $x$ to make explicit to which component or aspect of $x$ the text refers. For the genitive pattern \textit{$x$'(s) $y$}, we select insertions including an apostrophe and a noun $y$ that is in a dependency relation \texttt{nmod:poss} with a noun $x$. For the \textit{$y$ of $x$} pattern, we select insertions that consist of a noun $y$ and the token \textit{of} added right in front of a noun $x$, allowing for intervening determiners and adjectives.

\subsection{Constructing Clarifications}
\label{subsec:step2}

We produce a set of possible clarifications for each instance as follows: First, we generate the top-100 fillers in place of an observed insertion using 
language modeling. Second, we select a subset of potentially suitable clarifications by filtering and clustering the top-100.  

\paragraph{Filler generation.} For the implicit references, we take the 
top-100 generated clarifications from \newcite{anthonio-roth-2021-resolving}. For the other phenomena, we 
generate alternative clarifications automatically
using 
the same approach as \newcite{anthonio-roth-2021-resolving}. That is, we feed the original sentence $s$ with the surrounding sentences from the same 
paragraph to a language model. We then compute the top-$100$ completions for the token position(s) where an insertion was added in the revised sentence.
We use BERT \cite{devlin-etal-2019-bert} instead of GPT \cite{Radford:2018:gpt} to generate the clarifications, as the required insertions consist of only one token and BERT makes it possible to also consider follow-up context directly.
The BERT checkpoint \texttt{bert-base-uncased}\texttt{bert-base-uncased}\footnote{
We also tried \texttt{bert-base-cased} in preliminary experiments but observed no improvements.}
in \texttt{Transformers} \cite{wolf-etal-2020-transformers} was used without additional pre-training.

\paragraph{Filler selection.}
From the top-$100$ clarifications provided by the language model, we select four fillers with the goal of producing a semantically diverse set of clarifications.
First, we remove unsuitable fillers from the top-100, including 
cases
that only consist of digits or non-alphanumerical characters and fillers that do not have the right part of speech based on \texttt{Stanza} (retaining only \textit{NOUN} for fused heads and metonymy and \textit{NN} for noun compounds to exclude plural nouns).

For all instances with $\ge 4$ candidate fillers, we select the observed insertion from the revised sentence as one filler. To select semantically different fillers as alternate candidates, we apply $k$-means clustering with $k = 4$ to the remaining candidates, using the algorithm 
by \newcite{Elkan-03} as implemented in \texttt{sklearn} \cite{scikit-learn}. We obtain vector representations for clustering from BERT (\texttt{bert-base-uncased}) by averaging over the last hidden state for all tokens in a filler. 
After clustering,
we select the fillers closest to the four cluster centroids based on cosine similarity.

\subsection{Plausibility Annotation}
\label{subsec:step3}

\paragraph{Task.} After selecting fillers for each sentence, we collect plausibility judgements on Amazon Mechanical Turk for our train set (19,975 instances, i.e.~3995 sentences with 1 human and 4 generated fillers each\footnote{1000 each for noun compounds and metonymy, 996 for implicit references and 999 for fused heads.}), development and test sets (2,500 instances each, i.e.,~125 sentences per phenomenon with 5 fillers per sentence). 
Each clarification in the training set is annotated by 2 crowdworkers. For the development and test set, we collected annotations from 4 crowdworkers to ensure a consistently high quality.
In each annotation task, we ask participants to indicate on a scale from 1 to 5 whether the clarification made sense in the given how-to-guide. 
A screenshot of the interface for our Human Intelligence Task (HIT) is provided in Appendix~\ref{sec:appendix2}.

\begin{table}[!tbp]
\centering
\begin{tabular}{@{}l@{}l@{}l@{}l@{}}
\toprule
\textbf{}            
& \multicolumn{1}{l}{\textbf{Train}} 
& \multicolumn{1}{l}{\textbf{Dev}}
& \multicolumn{1}{l@{~}}{\textbf{Test}} \\ \midrule
\small \textsc{implausible} & ~~5,474 (27\%) 
& ~~~982 (39\%) 
& ~~~858 (34\%) 
\\
\small \textsc{neutral}     & 
~~7,162 (36\%) 
& ~~~602 (24\%) 
& ~~~672 (27\%) 
\\
\small \textsc{plausible}   & 
~~7,339 (37\%) 
& ~~~916 (37\%) 
& ~~~970 (39\%) 
\\
\midrule
\textbf{Total} & 19,975          & 2,500         & 2,500          \\ \bottomrule
\end{tabular}
\caption{Distribution of class labels in our training, development and test sets.}
\label{tab:splits}
\end{table}

\begin{table*}[!ht]
    \centering
    \begin{tabular}{ccll}
    \toprule
    Team & Model type & Pre-trained model components & Additional comments \\
    \midrule
    X-PuDu & ensemble & DeBERTa, ERNIE, XLM-R & pattern-aware, multi-loss\\
    HW-TSC & ensemble & DeBERTa, RoBERTa, S-BERT & incl.~unsupervised model\\
    PALI & ensemble & DeBERTa, RoBERTa, XLM-R & pattern-aware, multi-loss\\
    Nowruz & Transformer & T5 & ordinal regression, multi-loss\\
    JBNU-CCLab & ensemble & DeBERTa & $-$\\
    DuluthNLP & Transformer & ELECTRA & class weighting \\
    Stanford MLab & Transformer & ELECTRA & $-$\\
    niksss & Transformer & BERT & $-$\\    
    \bottomrule
    \end{tabular}
    \caption{Summary of the best models on the test set according to the submitted system descriptions.}
    \label{tab:systems}
\end{table*}

\paragraph{Qualifications.} We use several qualifications to increase the annotation quality. First, we require participants to be located in the United States or in the United Kingdom, to increase the chance that the participants are native speakers of English. Secondly, participants need to have a HIT approval rate $\ge 95\%$ and their number of approved HITS has to be $\ge 1000$. Finally, annotators are required to pass a qualification test in which they are asked to judge a list of clearly plausible and implausible cases that were pre-selected unanimously by the authors.

\paragraph{Class labels.} For Task~1 (classification),
we average over the real-valued judgements collected for a clarification and map this plausibility score to one of the three classes labels. Specifically, we label clarifications with an average score $\leq 2.5$ as \textsc{implausible}, clarifications with a score $\geq 4.0$ as \textsc{plausible}, and all clarifications between these thresholds as \textsc{neutral}. 
The thresholds have been selected based on manual inspection of the data and mathematical considerations: in particular, the threshold for \textsc{plausible} requires scores to be substantially above average (in case of two judgements, $\geq$3\&5 or $\geq$4\&4), whereas the \textsc{implausible} threshold allows for a slightly wider range of judgements. The \textsc{neutral} label covers cases that received inconclusive individual scores as well as cases of disagreement (e.g.~3\&3 as well as 2\&5).

\paragraph{Statistics.} 
We show the frequency distribution of the labels in the train, development and test set in Table~\ref{tab:splits}. 
It is noteworthy that development and test set proportionally includes fewer \textsc{neutral} and more \textsc{implausible} clarifications than the training set. Presumably, this is because we increased the number of qualification questions from 4 to 6 after collecting the training data to ensure the quality of the evaluation data.

Since we are particularly interested in cases with multiple plausible clarifications, we also compute the average number of \textsc{plausible} clarifications per sentence $s$, which we found to be 1.84, 1.87 and 1.84 in the training, development and test set, respectively. This means that, on average, each annotated sentence in the dataset has between 1 and 2 clarifications that the annotators rated as plausible.

\section{Participants}
\label{sec:participants}

A total of 21 users participated in the CodaLab competition set up for the shared task and 8 teams submitted system description papers. An overview of the best model by each team is shown in Table~\ref{tab:systems}.\footnote{A table with the official results of the CodaLab competition, including participants who did not submit system descriptions, is shown in Appendix~\ref{sec:appendix}.} We observe that all systems are based on Transformer architectures, using one or more of the following pre-trained models: BERT \cite{devlin-etal-2019-bert}, DeBERTa \cite{he:2020:deberta}, ELECTRA \cite{clark:2019:electra}, ERNIE \cite{sun:2019:ernie}, RoBERTa \cite{liu:2019:roberta}, S-BERT \cite{reimers-gurevych-2019-sentence}, T5 \cite{raffel:2020:t5}, XLM-R \cite{conneau-etal-2020-unsupervised}. 

In addition to fine-tuning a single or multiple Transformer models in an ensemble, some teams have taken additional steps to adapt their system to the task. We summarize some of these steps below.

\paragraph{Consideration of phenomena.} At least two teams took into account that the data set consists of four phenomena that were identified using different patterns (\textit{pattern-aware}): PALI used the phenomenon description that applies to a classification instance as additional model input; X-PuDu developed an ensemble architecture that consists of different individual models and hyperparamters for each phenomenon.

\paragraph{Adapted loss functions.} Several teams adapted the loss functions of their models to better account for various properties of the task. This includes the use of classification and regression based loss functions in a multi-task learning set-up (\textit{multi-loss}) as well as the use of specific loss functions that consider the ordinal nature of labels (\textit{ordinal regression}) or differences in label distributions (\textit{class weighting}) in the classification task.

\paragraph{Unsupervised components.} Given the similarity of our task to general cloze tasks, several teams experimented with models that were merely self-supervised and not fine-tuned on task-specific training data. In case of one team, HW-TSC, such an unsupervised component is also part of the ensemble model that produced the best results.



\section{Results and Discussion}
\label{sec:results}

The results for Task~1 and~2 are shown in Table~\ref{tab:task1results} and~\ref{tab:task2results}, respectively. We focus our discussion on Task~1: Classification, as the participants of Task~2 form only a subset of the Task~1 participants and the system results rank, with exception of the last two teams, in the same order. In addition to showing results by participants, we also provide a human upper bound as well as results by our own BERT-based baseline model. The upper bound was computed as the accuracy over all individual annotations when compared against the  (averaged) class label of each test instance.

The human upper bound has an accuracy of 79.4\%, indicating that the task is challenging and potentially involves a number of disagreements. The winning team of the competition, X-PuDu, achieves an accuracy of 68.9\%, only 10.5 percentage points below the human upper bound. The results of all teams lies substantially above a naive majority class baseline of 39\%. All teams but one also outperform our BERT-based baseline, which is a linear classification model based on the checkpoint provided by the Transformer library \cite{wolf-etal-2020-transformers} and fine-tuned on our training data.

\subsection{Findings by Participants}
\label{subsec:Findings}

In the following, we briefly summarize a couple of findings by task participants. More details can be found in the individual task description papers.

\begin{table}
    \centering
    \begin{tabular}{clc}
    \toprule
    Rank & Team & Accuracy\\
    \midrule
    $-$ & Human (upper bound) & 79.4\%\\
    \midrule
    1 & X-PuDu & 68.9\% \\
    2 & HW-TSC & 66.1\% \\
    3 & PALI   & 65.4\% \\
    4 & Nowruz & 62.4\% \\
    5 & JBNU-CCLab & 61.4\% \\
    6 & DuluthNLP & 53.3\% \\
    7 & Stanford MLab & 46.6\% \\
    8 & niksss & 44.2\% \\
    \midrule
    $-$ & BERT (baseline) & 45.7\% \\
 \bottomrule
    \end{tabular}
    \caption{Results for Task~1 (classification).}
    \label{tab:task1results}
\end{table}

\paragraph{Different phenomena.} The winning team, X-PuDu, found that different hyperparameters worked best depending on the phenomenon/extraction pattern. Based on this finding, different individual models were trained and combined in an ensemble.

\paragraph{Label distribution.} Some teams, including DuluthNLP, noticed performance issues related to the distribution of labels in the development data. As a dedicated solution, DuluthNLP uses a decreased weight for the \textsc{neutral} label in the loss function.

\begin{table}
    \centering
    \begin{tabular}{clc}
    \toprule
    Rank & Team & Spearman's $\rho$\\
    \midrule
    1 & X-PuDu & 0.807 \\
    2 & PALI & 0.785 \\
    3 & HW-TSC & 0.774 \\
    4 & Nowruz & 0.707 \\
    5 & niksss & 0.252 \\
    6 & Stanford MLab & 0.194 \\
    \bottomrule
    \end{tabular}
    \caption{Results for Task~2 (ranking).}
    \label{tab:task2results}
\end{table}

\paragraph{\textsc{neutral} label.} Team JBNU-CCLab reported that the \textsc{neutral} label is generally difficult to distinguish from other labels by different models. An underlying problem could be that the label represents instances that are seen as somewhat plausible by multiple annotators as well as instances that are seen as plausible by some annotators and implausible by others (see Section~\ref{sec:data}).

\paragraph{Noisy data.} Team HW-TSC found that isolated training instances have the label \textsc{neutral} rather than \textsc{plausible}, even though the respective filler represents a human insertion (i.e., the filler can be found in the final version of the text in wikiHow). As the results of our human upper bound in Table~\ref{tab:task1results} show, this is partly because the right label is sometimes not clear cut even for humans. We discuss this aspect in more detail in the next section.

\subsection{Additional Evaluations}
\label{subsec:analysis}

We perform two additional evaluations to assess the impact of the \textsc{neutral} label on system performance and to investigate the possibility of identifying whether multiple plausible clarifications exist by aggregating the predictions regarding individual clarifications.

\begin{table}
    \centering
    \begin{tabular}{clcc}
    \toprule
    Rank & Team & F1 (all) & F1 (w/o \textsc{n})\\
    \midrule
    1 & X-PuDu       & 0.689 & 0.773\\ 
    2 & HW-TSC       & 0.661 & 0.749\\
    3 & PALI         & 0.654 & 0.749\\ 
    4 & Nowruz       & 0.624 & 0.714\\
    5 & JBNU-CCLab   & 0.551 & 0.627\\
    6 & DuluthNLP    & 0.533 & 0.608\\
    7 & Stanford MLab& 0.466 & 0.514\\
    8 & niksss       & 0.442 & 0.494\\
 \bottomrule
    \end{tabular}
    \caption{Classification results with/without \textsc{neutral}.}
    \label{tab:noneutral}
\end{table}

\begin{table}
    \centering
    \begin{tabular}{clc}
    \toprule
    Rank & Team & Accuracy (\#\textsc{p}$\ge$2)\\
    \midrule
    1 & X-PuDu       & 75.2\% \\ 
    2 & HW-TSC       & 73.2\% \\
    3 & PALI         & 72.6\% \\ 
    4 & Nowruz       & 71.6\% \\
    5 & JBNU-CCLab   & 63.6\% \\
    6 & DuluthNLP    & 62.6\% \\
    7 & Stanford MLab& 54.8\% \\
    8 & niksss       & 60.0\% \\
 \bottomrule
    \end{tabular}
    \caption{Results for identifying contexts with multiple plausible fillers, based on individual model predictions.}
    \label{tab:multiple}
\end{table}

\begin{table*}[!h]
    \centering
    \begin{tabular}{@{~}l@{~}c@{~}p{9.7cm}l@{~}}
    \toprule
        Correct & \#\textsc{P}$\ge$2 & \multicolumn{1}{c}{Text} & Fillers\\
    \midrule
         \multirow{6}{1.1cm}{8 (all)}  & \multirow{3}{0.5cm}{\cmark} & Galette des rois—or ``King Cake'' in English—is traditionally made to celebrate the \underline{~~~~$\emptyset$~~~~} of Epiphany. Especially popular in France during the Christmas season, it is enjoyed elsewhere too. & 
         \multirow{3}{3.2cm}{\cmark ~holidays \cmark ~Feast $\textbf{?}$~hours~$\textbf{?}$~celebration $\textbf{?}$~proclamation} \\
         
         \cmidrule{3-4}
         
         & \multirow{3}{0.5cm}{\xmark} & Let the \underline{~~~~$\emptyset$~~~~} of shoes air dry. 	You can put them in front of a dehumidifier, a fan, or an open window, but avoid putting them in front of any type of heat source. & 
         \multirow{3}{3.2cm}{\cmark ~pair $\textbf{?}$ pile \quad \xmark ~shoes \xmark ~color \xmark ~end} \\
         
         \midrule
         
         \multirow{6}{1.3cm}{0 (none)} & \multirow{3}{0.5cm}{\cmark} & If you want a smoother surface, try a \underline{~~~~$\emptyset$~~~~} of paper with a higher amount of grains, if you want a faster job but a rougher surface try a paper with a lower amount of grains. & 
         \multirow{3}{3.2cm}{\cmark ~thickness \cmark ~piece $\textbf{?}$~fabric $\textbf{?}$~product \xmark ~pile} \\
         
        \cmidrule{3-4}
        
         & \multirow{3}{0.5cm}{\xmark} & Your cucumber plant will also grow thin, light green shoots that help the plant grasp onto a surface and grow vertically. These \underline{~~~~$\emptyset$~~~~} grow immediately next to the suckers.& 
        \multirow{3}{3.2cm}{\cmark ~shoots $\textbf{?}$~fibers $\textbf{?}$~tendrils $\textbf{?}$~foliage \xmark ~bushes} \\
    \bottomrule
    \end{tabular}
    \caption{Examples of difficult and easy instances, selected based on how many systems classified them correctly.}
    \label{tab:easyhard}
\end{table*}

\paragraph{Excluding \textsc{neutral}.} For the evaluation without the \textsc{neutral} label, we calculate micro-averaged precision, recall and F$_1$-scores for the two labels \textsc{plausible} and \textsc{implausible}. The results in terms of F$_1$-score are shown in Table~\ref{tab:noneutral}. The results indicate that all systems perform substantially better in the evaluation setting that ignores \textsc{neutral} labels. The ranking is identical to the ranking in the evaluation including all labels. Considering only the \textsc{plausible} and \textsc{implausible}, Team X-PuDu achieves the highest micro-averaged F$_1$-score of 0.773. In the cases where their system predicts a non-\textsc{neutral} label, it is correct in 72.7\% of cases (precision), and 82.5\% of all non-\textsc{neutral} instances in the data received the correct prediction (recall). 

\paragraph{Multiple clarifications.} In our final evaluation, we examine whether system predictions can also be used to determine whether multiple plausible clarifications for a given context exist. For this, we consider the labels of each individual clarification and compare system outputs and annotations in terms of whether two or more clarifications for a cloze and its context received the label \textsc{plausible}. We show the result of this evaluation in terms of accuracy for each team in Table \ref{tab:multiple}. Apart from the last two places, the teams rank in the same order as in the other evaluations. The best performing team, X-PuDu, correctly predicts whether two or more plausible clarifications exist for 75.2\% of all cases. 
Table~\ref{tab:easyhard} shows examples that were correctly classified by all or none of the systems. 

\section{Conclusion}
\label{sec:conclusion}

In this paper, we presented the task, data, participating systems, and results of the shared task on clarifying implicit and underspecified phrases in instructional texts. Our motivation for this task was to explore the possibility of testing different clarifications for plausibility. In particular, we were concerned with the question of whether two or more clarifications can be plausible and whether such cases can be detected automatically. To create a suitable dataset, we worked with and identified a set of revisions with manual clarifications, automatically generated possible alternatives, and then collected human plausibility ratings.

In total, 21 users participated in our shared task. We summarized the systems and results of 8 teams that submitted descriptions of their systems. The best systems from each group have in common that they are based on Transformer architectures or combine them in an ensemble. The best system achieved 68.9\% accuracy, only 10.5 percentage points below a human upper bound. In additional evaluations, we have shown that an accuracy of up to 75.2\% is achieved with respect to the detection of multiple plausible clarifications.

The results show that the presented task is a difficult one, but that many cases can already be modeled well by current state-of-the-art methods. There is further room for improvement with respect to both the data set and models: with respect to the data, it should be noted that the training set with less than 20k instances is relatively small and that there are many instances with a underspecified \textsc{neutral} label (36\%). On the model side, we found that the participating teams make complementary contributions that may allow for additional improvements in combination.

One shortcoming of the task as presented and performed is that we only considered four forms of clarifications related to referring expressions. In addition, clarifications were assessed individually and judgements by different annotators were aggregated. In the long term, we believe that more forms of clarifications as well as individual differences regarding their plausibility need to be considered. Finally, future work will have to investigate under which circumstances multiple different clarifications are actually incompatible and can thus reveal potential sources of misunderstanding.


\section*{Acknowledgements}

The research presented in this paper was funded by the DFG Emmy Noether program (RO 4848/2-1).


\bibliography{anthology,custom}
\bibliographystyle{acl_natbib}

\appendix

\section{Annotation Interface}
\label{sec:appendix2}

\begin{figure}[h!]
    \centering
    \includegraphics[scale=0.35]{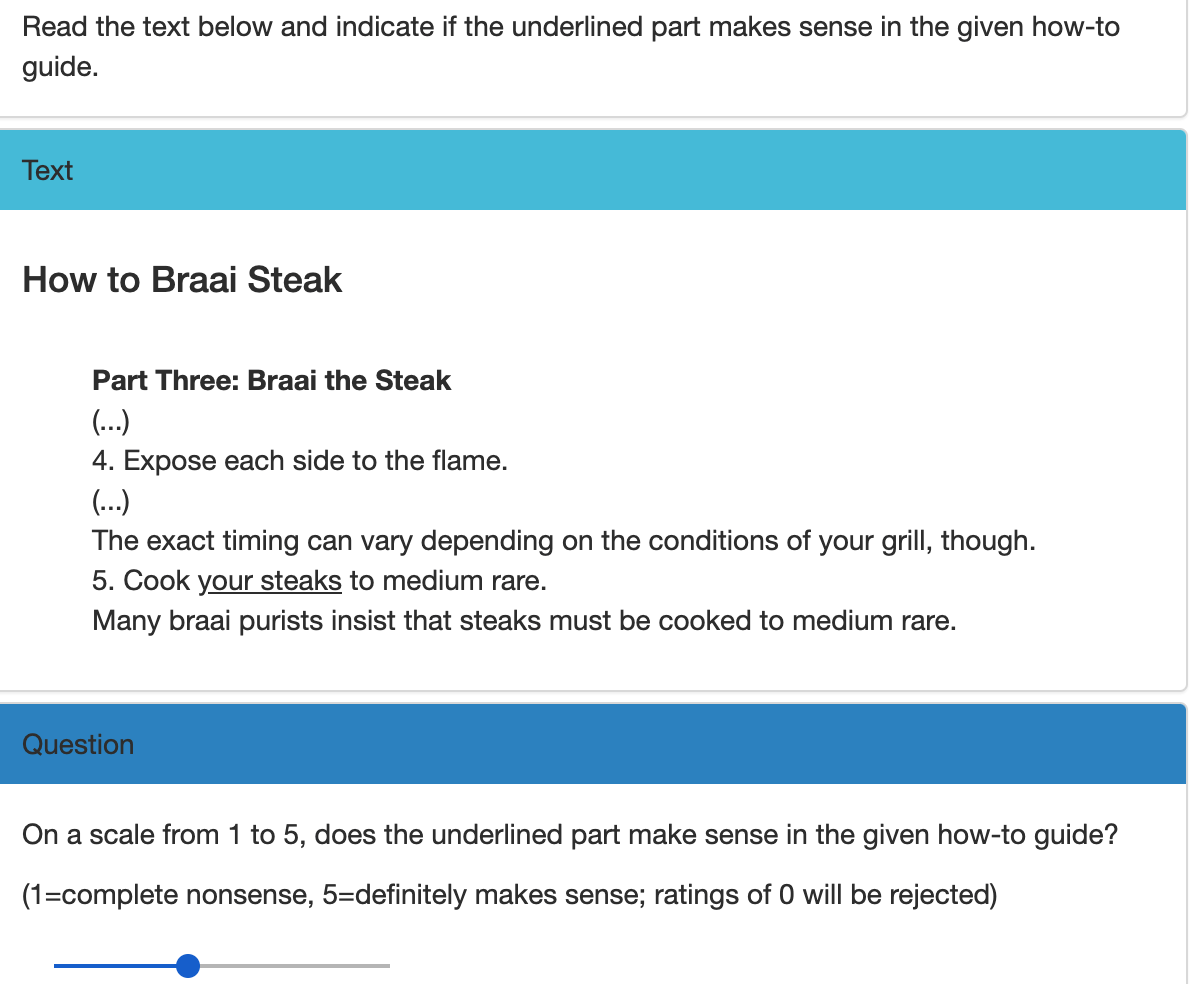}
    \caption{Interface for collecting annotations.}
    \label{fig:annotation-interface}
\end{figure}

The annotation interface for our crowdsourcing task is depicted in Figure \ref{fig:annotation-interface}.
Annotators see and have to rate a single underlined clarification in its context.

\section{CodaLab Leaderboard}
\label{sec:appendix}

In the main part of the paper, we only list results of participants who provided a description of their system(s) for the shared task. Table~\ref{tab:allresults} shows a complete set of user names and results of the participants in the CodaLab competition, including users who did not submit a system description. 

\newpage

\begin{table}[!th]
    \small
    \begin{tabular}{@{}llcc@{}}
    \toprule
    Team name & User name & acc. & $\rho$\\
    \midrule
X-PuDu & tt123        & 0.689 & 0.807 \\
HW-TSC & Yinglu\_Li    & 0.661 & 0.774 \\
$-$ & tiantaijian	& 0.661 & 0.763 \\
$-$ & fanxiaoxing	& 0.656 & $-$ \\
PALI & stce         & 0.654 & 0.785 \\
$-$ & hudou	    & 0.641 & $-$ \\
$-$ & huangwkk     & 0.631 & 0.774 \\
Nowruz & mohammadmahdinoori & 0.624 & 0.707 \\
JBNU-CCLab & OrangeAvocado& 0.614 & $-$ \\
$-$ & CitizenTano	& 0.595 & $-$ \\
$-$ & huawei\_zhangmin & 0.589 & 0.640 \\
$-$ & parkwonjae	& 0.554 & $-$ \\
$-$ & lith	        & 0.537 & 0.600 \\
$-$ & ywzhang\_cr	& 0.537 & 0.600 \\
DuluthNLP & Sakrah	    & 0.533 & $-$ \\
Stanford MLab & patrickliu2011 & 0.466 & 0.194 \\
$-$ & Autism\_PAFC	& 0.461 & $-$ \\
$-$ & SelinaIW     & 0.456 & $-$ \\
niksss & niksss	    & 0.442 & 0.252 \\
$-$ & andrei.manea & 0.418 & -0.109 \\
$-$ & tanigaki	    & 0.395 & 0.415 \\
    \bottomrule
    \end{tabular}
    \caption{Oveview of results, including user submissions without a shared task system description.}
    \label{tab:allresults}
\end{table}

\end{document}